\documentclass{article}

\usepackage{arxiv}

\usepackage[utf8]{inputenc} 
\usepackage[T1]{fontenc}    
\usepackage[hidelinks]{hyperref}       
\usepackage{url}            
\usepackage{booktabs}       

\usepackage{mathtools}
\usepackage{amsfonts}       
\usepackage{nicefrac}       
\usepackage{microtype}      
\usepackage{lipsum}		

\usepackage[round]{natbib}

\usepackage{graphicx}
\usepackage{doi}
\usepackage{amssymb}
\usepackage{pifont}
\usepackage{xcolor}
\usepackage{caption}
\usepackage{multirow}
\usepackage{tabularx}
\usepackage{multicol}
\usepackage{cleveref}
\usepackage{stfloats}
\usepackage[caption=false,font=normalsize,labelfont=sf,textfont=sf]{subfig}
\usepackage{algorithmic}
\usepackage{array}

\title{A Survey on Knowledge Editing of Neural Networks}

\date{} 					

\author{%
    Vittorio Mazzia\\
    Alexa AI, Amazon\\
    \texttt{vmazzia@amazon.com} \\
    \And
    Alessandro Pedrani\\
    Alexa AI, Amazon\\
    \texttt{pedrana@amazon.com} \\
    \And
    Andrea Caciolai\\
    Alexa AI, Amazon\\
    \texttt{andccl@amazon.com} \\
    \AND
    Kay Rottmann\\
    Alexa AI, Amazon\\
    \texttt{krrottm@amazon.com} \\
    \And
    Davide Bernardi\\
    Alexa AI, Amazon\\
    \texttt{dvdbe@amazon.com} \\
}



\hypersetup{
    colorlinks=true,
    linkcolor=blue,
    urlcolor=blue,
    citecolor=blue
}

\usepackage{cleveref}

\definecolor{darkgreen}{RGB}{0,100,0} 
\newcommand{\greencheck}{\textcolor{darkgreen}{\checkmark}}
\newcommand{\redxmark}{\textcolor{red}{\ding{55}}}

\begin{document}
\maketitle

\begin{abstract}
Deep neural networks are becoming increasingly pervasive in academia and industry, matching and surpassing human performance on a wide variety of fields and related tasks. However, just as humans, even the largest artificial neural networks make mistakes, and once-correct predictions can become invalid as the world progresses in time. Augmenting datasets with samples that account for mistakes or up-to-date information has become a common workaround in practical applications. However, the well-known phenomenon of catastrophic forgetting poses a challenge in achieving precise changes in the implicitly memorized knowledge of neural network parameters, often requiring a full model re-training to achieve desired behaviors. That is expensive, unreliable, and incompatible with the current trend of large self-supervised pre-training, making it necessary to find more efficient and effective methods for adapting neural network models to changing data. To address this need, knowledge editing is emerging as a novel area of research that aims to enable reliable, data-efficient, and fast changes to a pre-trained target model, without affecting model behaviors on previously learned tasks. In this survey, we provide a brief review of this recent artificial intelligence field of research. We first introduce the problem of editing neural networks, formalize it in a common framework and differentiate it from more notorious branches of research such as continuous learning. Next, we provide a review of the most relevant knowledge editing approaches and datasets proposed so far, grouping works under four different families: regularization techniques, meta-learning, direct model editing, and architectural strategies. Finally, we outline some intersections with other fields of research and potential directions for future works.
\end{abstract}


\section{Introduction}
In stark contrast to artificial neural networks (ANN), \cite{cichon2015branch}, humans and other animals seem capable of learning and editing their knowledge continuously. 
Indeed, literature studies indicate that the mammalian brain could prevent catastrophic forgetting \cite{ratcliff1990connectionist} by safeguarding previously acquired knowledge, thereby reducing the plasticity of a proportion of synapses and ensuring their long-term stability \cite{benna2016computational, yang2009stably, cichon2015branch}. 
On the contrary, ANNs not only struggle to learn new tasks in a sequential fashion \cite{kirkpatrick2017overcoming}, but also edit acquired knowledge on the same data distribution and task \cite{huangtransformer}. 
Indeed, unlike conventional knowledge base systems that explicitly store knowledge, neural models implicitly memorize facts and tasks in their parameters, making it difficult to directly access and interpret their computation and memories \cite{voita2019bottom, belinkov2019analysis}. 
Making even minor modifications can lead to a decrease in performance on previously learnt tasks, or even cause the entire computation to fail due to the well-documented issue of catastrophic forgetting \cite{ratcliff1990connectionist}. 
Therefore, modifying their acquired knowledge is a challenging problem. 

Just as humans, ANNs make mistakes and as we trust them with increasingly important decisions, the cost of such mistakes grows ever higher \cite{sinitsineditable}. 
Therefore, given that mistakes are inevitable, it is crucial for deep learning practitioners to possess the ability to adjust model behaviors by promptly correcting errors as they surface. Currently, practical applications employing deep learning techniques have been relying on different workarounds to tackle this problem. 
In particular, a full re-training using datasets augmented with samples that account for the mistakes or up-to-date information is a common choice \cite{sinitsineditable}. The endeavor needed for fine-tuning atop pre-trained models \cite{sarzynska2021detecting, devlin2018bert, oquab2023dinov2, weiss2016survey} is frequently substantiated by the diminished dataset size and computational resources needed.
On the other hand, this is not always true and manually curated, deterministic mechanisms that overrule model predictions on problematic samples can be the preferred choice \cite{sinitsineditable}. 
However, while being simple, this second approach is fully localized and not robust to factor of variations of the input space (e.g., different viewpoint of the same object in computer vision or paraphrasing in natural language processing tasks). 
Furthermore, while these workarounds may provide a temporary solution, they can be expensive, unreliable, and incompatible with the current trend of large neural models \cite{zhao2023survey, chen2022pali}. 
Indeed, these large networks are typically deployed as static artifacts, whose behavior is difficult to modify during deployment without a costly full re-training \cite{lazaridou2021mind}. 
Thus, in all those cases, in order to adapt to changes in the environment, or to address instances of underfitting or overfitting in the original training data, it is desirable to have the ability to quickly make targeted updates to the model's behavior after it has been deployed \cite{de2021editing}.

To address this need, \textit{knowledge editing} methods have been recently proposed to efficiently change a model's behaviors without affecting previous performance on the same task \cite{sinitsineditable}. These approaches take inspiration from several fields of artificial intelligence research and range from simple fine-tuning with regularization methods \cite{zhu2020modifying} to meta-learning techniques that adopt hypernetwork models to learn how to update parameters \cite{de2021editing, mitchellfast}. Due to its recent appearance in the literature, \cite{sinitsineditable}, the field still lacks accordance in the taxonomy, naming convention, datasets, and target applications. Indeed, most of the works have been motivated by large language models (LLMs), \cite{zhao2023survey, brown2020language, soltan2022alexatm}, focusing mostly on tasks such as question answering (QA), \cite{levy2017zero}, machine translation (MT) \cite{de2021editing}, modifying knowledge graph embeddings \cite{cheng2024editing}, or even simpler NLP problems \cite{thorne2018fever}. However, it is also possible to find applications of knowledge editing to computer vision problems \cite{sinitsineditable}. Furthermore, its potential scope is poised to expand across diverse machine learning domains, encompassing areas such as medicine \cite{shehab2022machine} robotics \cite{soori2023artificial}, and precision agriculture \cite{sharma2020machine} in the future.

\subsection{Organization of the survey}
The objective of this survey is to provide a comprehensive review of existing literature on knowledge editing, formalizing the task and providing a categorization of the approaches into distinct families. To our knowledge, this is the first work to undertake such an effort, and we hope that it will facilitate future research in this increasingly important area of study. Indeed, the need for more formalization and organizational structure can already be seen by a recent study \cite{yao2023editing}, which attempts to benchmark and compare some knowledge editing methodologies specifically for LLMs editing. 

The rest of the survey is organized as follows. 
\ref{sec:overview} introduces the problem of knowledge editing, using previous works to formalize it under a common definition. 
\ref{sec:tasks-datasets} explores the tasks and datasets that are most commonly considered when solving the knowledge editing problem. 
\ref{sec:methodologies} provides an overview of the most relevant knowledge editing approaches in the literature, identifying four distinct families: \textit{regularization techniques}, \textit{meta-learning}, \textit{direct model editing}, and \textit{architectural strategies}. 
Finally, \ref{sec:conclusion} concludes the survey by discussing some intersection with knowledge editing and other fields of research, and outlining some possible future risks and directions.

\section{Overview}
\label{sec:overview}
This section begins by presenting an introduction to the concept of \textit{knowledge editing}, which is also referred to as \textit{model editing} in the literature. First, we review various definitions and interpretations of knowledge editing proposed by different works. Next, we establish a common definition of knowledge editing that generalizes to all existing works in the field.

\subsection{Background}
The concept of knowledge editing was first introduced in \cite{sinitsineditable}, which formalizes it as the task of correcting a model's mistake on a \textit{specific sample} while preserving the model's overall behavior, akin to continuous learning. 
Indeed, as specified by the authors, “The problem of efficient neural network patching differs from continual learning, (...) [because] is not sequential in nature. However, correction (...) of mislabeled samples must not affect its behavior on other samples, which is close to overcoming [the] catastrophic forgetting task.”. 
Therefore, \cite{sinitsineditable} define the problem as performing individual edits \textit{reliably} and \textit{efficiently}, not \textit{sequentially}, and on the same task learned by the target model without affecting its overall behavior (i.e., being \textit{local} to the edit). 
Concurrently, authors of \cite{zhu2020modifying} worked specifically on the task of modifying memories in Transformer models \cite{vaswani2017attention}, providing their own definition of knowledge editing. They expand the scope of the problem to a subset of knowledge, i.e., a \textit{batch of edits}. 
Similarly, several other studies have also formalized the problem of model editing, following similar steps to \cite{zhu2020modifying}. 
For instance, works by \cite{mitchellfast}, \cite{meng2022mass}, and \cite{mitchell2022memory} have defined the task as the process of performing individual or batch edits on a target model trained for a specific task. 
These studies emphasize the importance of ensuring that edits are resilient to factors of variation, i.e., that they are generalizable. While injecting individual changes is already a challenging and interesting task for the scientific community, multiple simultaneous general model edits represent a more realistic scenario that deserves further exploration.

More recently, \cite{hartvigsen2022aging} and \cite{huangtransformer} argued that the conventional "one-mistake-fixing scenario" does not accurately reflect the complexity of real-world knowledge editing challenges. As such, they proposed to extend the scope of knowledge editing to a sequential problem to facilitate the development of more practical editing methods. While their proposal only accounts for subsequent individual edits, considering multiple simultaneous and sequential edits represents a more general case where the number of edits varies at each step. Importantly, in the case of iterative model editing, it is desirable to respect not only the new editing task constraints but also the previous ones, which is closely related to the concept of continual learning. Nevertheless, it is crucial to highlight that while the new definition of knowledge editing acknowledges a sequential process, differing from continuous learning, its scope remains limited to the modification of knowledge of the initially learned task by the model. On the contrary, continuous learning operates without such constraints, researching for methodologies that allow the model to expand to new tasks and adapt dynamically to entirely new information.

\subsection{The knowledge editing problem}
\label{sec:knowlege_editing_problem}
\begin{figure*}[t]
    \centering
    \includegraphics[width=\textwidth]{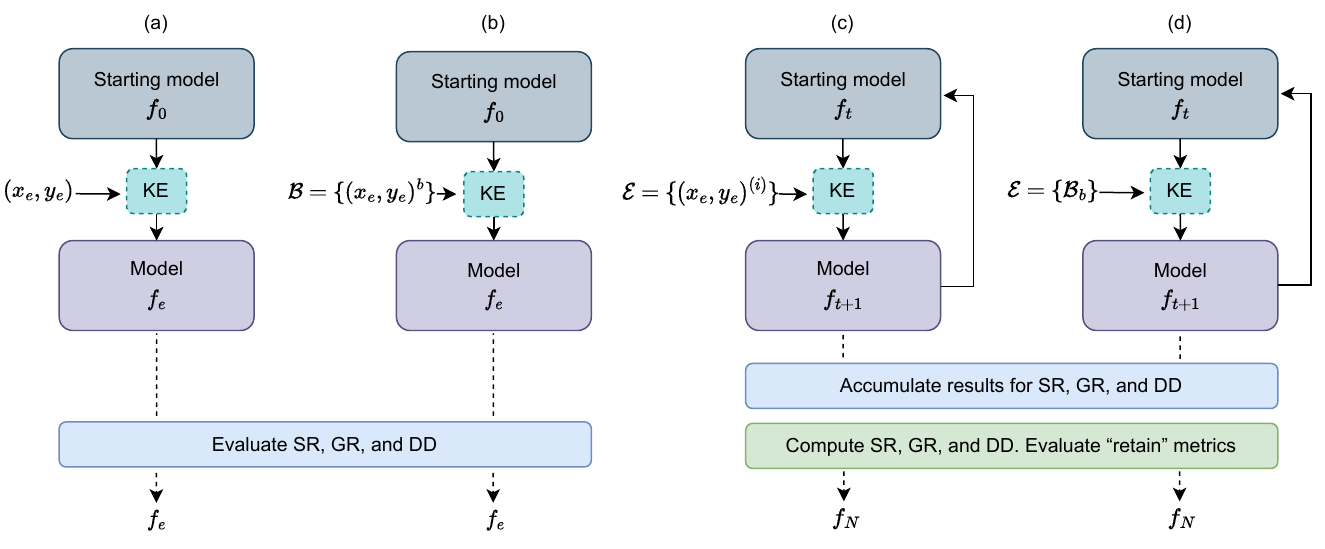}
    \caption{The knowledge editing problem has been firstly proposed as the task of modifying a model based on a set of individual pairs of edits, in a non-sequential manner (a), \cite{sinitsineditable}. Successive works extended the problem to batch of edits (b), sequential individual edits (c), and sequential batch of edits (d). Evaluation metrics are similar to all cases, as described in Section \ref{sec:evaluation_metrics}.}
    \label{fig:summary_ke}
\end{figure*}

To introduce the problem of knowledge editing, it is possible to leverage the terminology used to define a ``well-posed learning problem'' \cite{mitchell2007machine}: an algorithm is said to learn from experience $E$ with respect to some class of tasks $T$ and performance measure $P$, if its performance at tasks in $T$, as measured by $P$, improves with experience $E$. Then, we can say that knowledge editing is the problem of modifying the algorithm such that, given a representative set $S$ of instances of tasks in $T$, and a subset $S_e \subseteq S$ of \textit{edit} instances, its performance on $S_e$ improves as measured by $P$, while performance on all the other instances $S \setminus S_e$ remains unchanged. 

More practically, we can define the problem of knowledge editing as \textit{“the task of modifying a model based on a set of individual or batch edits}, $S_{e}$, \textit{pertaining to the same task known by the model, either in a sequential or non-sequential manner. The objective is to update the model's knowledge representation without significantly altering its original behavior or performance over} $S$ \textit{and being robust to different factor of variations of the edits.”}

More formally, let $\mathbb{X}, \mathbb{Y}$ be an input and output space, respectively. Let $f_{0}$ be a function that maps an input $x \in \mathbb{X}$ to an output $y \in \mathbb{Y}$, parametrized by $\theta_0 \in \Theta$, then
\begin{equation}
f_{0} \in \mathbb{F} \coloneqq  \left(\mathbb{X} \times \Theta \right) ^ \mathbb{Y}
\end{equation}
We use the subscript zero to indicate that this is the \textit{starting model}, i.e., the model we want to edit.
We define an \textit{edit pair} as an input-output pair $(x_{e}, y_{e}) \in \mathbb{X} \times \mathbb{Y}$, such that $f_{0}(x_e) \neq y_e$.
Then, in its simplest form, given an individual edit example pair $(x_{e}, y_{e})$, a knowledge editing (KE) methodology can be defined as follows
\begin{equation}
    \textrm{KE}: \mathbb{F} \times \mathbb{X} \times \mathbb{Y} \to \mathbb{F}
    \label{eq:ke-definition}
\end{equation}
i.e. a function that takes in input the starting model $f_0$ and the edit pair $(x_{e}, y_{e})$ to produce an \textit{edited model} $f_{e}$. If the edited model is such that $f_{e}(x_e) = y_e$, then we say that the edit was successful.
A KE approach can realize the transformation from $f_0$ to $f_{e}$ in different ways, and we identify four families of possible realizations: regularization, meta-learning, direct model editing and architectural strategies. We deep dive into more details into each of these families in \ref{sec:methodologies}.

\subsection{Taxonomy of edits}

Often, it is interesting to edit a model applying multiple edits at once, a sequence of edits, or sequences of batches of edits. 

The definition provided in section \ref{sec:knowlege_editing_problem} has been given for the simplest case, that we call a \textit{single non-successive} edit: we only want to change the model for one edit pair.
Conversely, for multiple successive edits we can formally define a list of edit requests as:
\begin{equation}
    \mathcal{E} = \{(x_{e}, y_{e})^{(i)}\hspace{0.2cm} \textrm{s.t.}\hspace{0.2cm} \forall i,j \hspace{0.2cm} x_e^i = x_e^j   \Rightarrow y_e^i = y_e^j\}
\end{equation}
where the logical constraint ensures that there are no conflicting requests, as suggested by \cite{meng2022mass}. 
Individual edits can also be grouped together and form $N$ batches of successive edits each with $B_e^{(i)}$ edit pairs, such as 
\begin{equation}
\mathcal{B}_e^{(i)} = \{(x_e, y_e)^{0}, \cdots, (x_e, y_e)^{B}\}^{(i)}\hspace{0.2cm} \textrm{s.t.}\hspace{0.2cm}  \mathcal{E} = \bigcup_{i=1}^{N} \mathcal{B}_e^{(i)}
\end{equation}
The difference between successive individual or batch of edits is that some KE methodologies can ingest an entire $\mathcal{B}_e^{(i)}$ and produce the resulting $f_{e}$ implementing all the given edits, while other methodologies can only consider one individual sample in sequence at a time.
In both cases (a sequence of individual edits is trivially a sequence of single-item batch edits), successive edits assume to work with a starting model $f_{t-1}$ and apply the $t$-th change as
\begin{equation}
    f_{t} = \textrm{KE}(f_{t-1}, \mathcal{B}_e^{(i)})
\end{equation}
proceeding iteratively, using $f_{t}$ as a starting model for the next edit. Figure \ref{fig:summary_ke} summarizes the four types of edits. Finally, as for individual edits, after every change, $f_{e}$ should not only satisfy $f_{e}(x_e)=y_e$, but a series of other properties as discussed in the next section.

\subsection{Editing properties}
\label{sec:editing-properties}
Based on the specific task learned by function $f$, various properties can be specifically designed. However, at a fundamental level, following \cite{sinitsineditable, huangtransformer}, knowledge editing should aim at satisfying four properties, that we define below. \\
\textbf{Property 1} - \textit{Reliability}: Given an edit pair $(x_{e}, y_{e})$, the edited model $f_{e}$ should output the desired edit:
    \begin{equation}
        \label{eq:reliability}
        f_{e}(x_e) = y_{e}
    \end{equation}
\textbf{Property 2} - \textit{Generality}: 
    The edited model $f_{e}$ should be able to generalize to \textit{similar} examples to the edit pair. 
    This can be formalized by defining an \textit{equivalence neighborhood} $N(x_e) \subset \mathbb{X}$ and requiring that the edited model $f_{e}$ satisfies:
    \begin{equation}
        \label{eq:generality}
        f_{e}(\tilde{x}_{e}) = y_{e} \qquad \forall \ x_{e} \in N(x_e)
    \end{equation}
\textbf{Property 3} - \textit{Locality}: 
    The edited model $f_{e}$ should not alter the output of examples that are not similar to the edit pair. This can be formalized by defining a \textit{locality set} 
\begin{align}
    L(x_e) = \{ &(x_{loc}, y_{loc})\in \mathbb{X} \times \mathbb{Y} \notag \\
    &\textrm{s.t. } x_{loc} \notin N(x_e)  \land f_{0}(x_{loc})=y_{loc}\}
\end{align}
    and require that the edited model $f_e$ satisfies:
    \begin{equation}
        f_{e}(x_{loc})=y_{loc} \qquad \forall \ (x_{loc}, y_{loc}) \in L(x_e) 
    \end{equation} 
\textbf{Property 4} - \textit{Efficiency}: The proposed knowledge editing methodology should aim to achieve efficient editing, characterized by low compute and space complexity.

It is worth noting that some works in the literature have defined these same properties, or a subset of them, with different names and notations. Specifically, reliability may also be referred to as \textit{efficacy}, locality as \textit{specificity}, and generality as \textit{paraphrase}, \cite{meng2022locating, de2021editing}.
Moreover, as previously mentioned, depending on the specific field of application, other properties and related evaluation metrics can arise, such as \textit{fluency} and \textit{consistency} \cite{meng2022locating} when editing Language Models. Finally, the last property, namely efficiency, tends to be disregarded in academic literature. However, it is one of the main reasons KE is appealing over a simpler re-training of the base model. Furthermore, it plays a pivotal role in performing comparative analyses with baseline models and showcasing the ability of a particular methodology to modify a neural network with a substantial number of parameters.
We encourage future works to consider (at least) all four properties when measuring the effectiveness of a proposed KE methodology.


\subsection{Evaluation metrics}
\label{sec:evaluation_metrics}
Given a knowledge editing algorithm KE, a list of edit requests $\mathcal{E}$, a starting model $f_{0}$ and a test set $\mathcal{D}_{test}$, it is possible to measure the extent to which the properties above hold.
We consider working with $N$ batches of successive edits, each comprised of $B$ individual edits, without loss of generality (as mentioned above, if $N=1$, we have successive individual edits). 
Moreover, if the KE methodology is unable to process all $B$ edits concurrently, it can apply each edit individually in a non-sequential manner.

Following \cite{huangtransformer}, we represent with $I$ the indicator function, and define the following metrics. \\
\textbf{Success Rate} (SR): It is used to evaluate reliability, and it is simply the proportion of edits for which the methodology succeeds in changing the knowledge of a starting model $f_{t}$.
\begin{align}
    \label{eq:success_rate}
    \textrm{SR} &= \frac{1}{N}\frac{1}{B}\sum_{n=1}^{N}
    \sum_{b=1}^{B}I(f_{n,B}(x_{e;n,b}) = y_{e;n,b}) \nonumber \\
    &\quad \text{s.t. }  \bigcup_{n = 1}^N \bigcup_{b=1}^B (x_{e}, y_{e})_{n,b} = \mathcal{E}
\end{align}
In the case of non-sequential individual edits, $f_{n,B}=f_{n,b}$. Moreover, Eq. \ref{eq:success_rate} provides an overall value after $N$ successive edits, but it can be of interest to measure SR every $n$ edits, tracking changes over the sequence. \\
\textbf{Generalization Rate} (GR): It is used to evaluate generality, testing the post-edit model $f_{e}$, on the equivalence neighborhood set $N(x_{e;n,b}, y_{e;n,b})$, where $(x_{e;n,b}, y_{e;n,b})$ is the $n$-th batch, $b$-th edit pair. GR can be written as,
\begin{align}
\label{eq:generalization_rate}
\textrm{GR} &= \frac{1}{N}\frac{1}{B}\frac{1}{N_{b}}\sum_{n=1}^{N}
\sum_{b=1}^{B}\sum_{i=1}^{N_{b}}I(f_{n,B}(\tilde{x}_{e;n,b,i}) = \tilde{y}_{e;n,b,i}) \nonumber \\
&\qquad \text{s.t. } \forall n, b \ \bigcup_{i = 1}^{N_b} (\tilde{x}_{e}, \tilde{y}_{e})_{n,b,i} \subseteq N(x_{e}, y_{e})_{n, b}
\end{align}
where $N_b$ is the number of equivalent samples of the $b$-th edit pair.
Following \cite{mitchellfast}, we can also define \textbf{Edit Success} (ES) to summarize both SR and GR. It can be computed as the average accuracy of the edited model $f_{e}$ on the edit input(s), as well as inputs drawn from the equivalence neighborhood(s), that is,
\begin{equation}
\begin{aligned}
\textrm{ES} = & \frac{1}{N}\frac{1}{B} \sum_{n=1}^{N}
\sum_{b=1}^{B} \Bigl( I(f_{n,B}(x_{e;n,b}) = y_{e;n,b}) \\
& +  \sum_{i=1}^{N_{b}} \frac{1}{N_{b}} I(f_{n,B}(x_{e;n,b,i}) = y_{e;n,b,i}) \Bigr)
\end{aligned}
\end{equation}
where the same conditions for SR and GR hold and have been omitted for brevity. \\
\textbf{Drawdown} (DD): It is used to evaluate locality, and it is defined as the performance degradation of the edited model over $\mathcal{D}_{test}$. It is computed using the final edited model, $f_{N}$, that in case of successive edits is the result of $N$ steps.
\begin{equation}
    \textrm{DD} = 1 - \frac{\sum_{(x,y)\in \mathcal{D}_{test}}I(f_{N}(x)=y)}{\sum_{(x,y)\in \mathcal{D}_{test}}I(f_{0}(x)=y)}
\end{equation}

Finally, as suggested by \cite{huangtransformer} in the case of multiple successive edits, it is also important to evaluate SR and GR using the final model $f_{N}$, in order to assess how past edits are retained. Therefore, it is possible to define three additional metrics, \textbf{Success Retain Rate} (SRR), \textbf{Generalization Retain Rate} (GRR), and \textbf{Edit Success Retain} (ESR), simply using in Eq. \ref{eq:success_rate} and \ref{eq:generalization_rate}, $f_N$ instead of $f_{n,B}$.

\section{Tasks and Datasets}
\label{sec:tasks-datasets}
The formalization of the knowledge editing problem provided in the previous section is general, and many applications of knowledge editing to different tasks encompassing various fields can be formulated within that framework.
The brief, but rich history of the field has so far seen applications mainly to two broad fields: Computer Vision and Natural Language Processing. Indeed, \cite{sinitsineditable} provides experimental results on image classification and machine translation, and almost all the works that come after (and even before \cite{garnelo2018conditional,kirkpatrick2017overcoming}) demonstrate the effectiveness of their proposed approaches in one or more applications in these two fields. 

Even though the knowledge editing framework can be defined independently of the target domain and task, each specific application has its unique challenges and intricacies, which we explore in this section. \ref{sec:tasks-datasets:cv} covers the most common tasks in the Computer Vision domain, as well as the datasets on which the tasks are usually addressed, and how the knowledge editing problem can be instantiated in this context. Section \ref{sec:tasks-datasets:nlp} provides a similar overview for applications to the Natural Language Processing domain. Finally, \ref{sec:tasks-datasets:misc} describes tasks and datasets that do not strictly fit in any of the two domains above. 
 
\subsection{Computer Vision}
\label{sec:tasks-datasets:cv}
Computer Vision is a broad field with a long history, which generally attempts to extract meaningful representations from visual media to derive an understanding of the represented scenes \cite{szeliski2022computer}. Over the last years, deep learning methods have been shown to outperform previous state-of-the-art techniques in several applications in the field \cite{voulodimos2018deep}, and while more ``traditional'' techniques are still relevant for some applications, neural networks-based approaches have become the de facto standard for many others \cite{o2020deep}. Due to the importance and breadth of the field, and the relevance of neural networks therein, knowledge editing literature has found fertile grounds in Computer Vision, and has so far gravitated towards two primary tasks: Image Classification and Image Completion. A number of datasets are customarily used to test approaches to solve these tasks. They vary in terms of number of examples, classes and channels, as well as resolution of the representative images; Table \ref{tab:CVDatasets} provides an overview of the most commonly used ones.

\paragraph{Image Classification} The task of image classification is straightforward, we wish to label a complete image (or predetermined portion) with its most likely semantic category, e.g., horse, cat, or car \cite{szeliski2022computer}. In this context, an example is an image and its semantic label. The image, of predefined dimension (or resolution), is encoded as a 3D tensor $x^{(i)} \in \mathbb{R}^{W \times H \times C}$, where $W$ is the width of the image, $H$ the height of the image, and $C$ the number of channels, depending on whether the image is grayscale (1) or RGB (3) or something else (e.g., RGBD \cite{firman2016rgbd} with an additional depth channel). The editing task is then often formulated by artificially corrupting a subset of either the images or the labels. The latter is the more prevalent, and usually involves training models on a subset of a dataset and subsequently introducing random label shuffling within a withheld set to create an edit set ${(x_e^{(i)}, y_e^{(i)})_{i=1}^N}$, where originally $y^{(i)} \neq y_e^{(i)}$. Other works, such as \cite{sotoudeh2021provable}, introduce e.g. motion blur or fog \cite{mu2019mnist} to corrupt a subset of the images, creating an edit set where originally $x^{(i)} \neq x_e^{(i)}$.

Several datasets support knowledge editing experiments for this task, and a distinction is often made among ``toy'' and ``large-scale'' datasets. Well-known datasets like MNIST \cite{lecun2010mnist} and CIFAR-10 \cite{Krizhevsky2009LearningML}, which are widely used in the literature, are frequently employed for experimentation and belong to the former category. For more challenging and realistic scenarios, researchers turn to larger scale datasets, of which the most popular is surely the extensive ImageNet Database \cite{5206848}, which now encompasses over 10 million labeled images, spanning more than 20,000 object categories. Specifically, studies such as \cite{sinitsineditable} and \cite{lee2019overcoming} explore datasets derived from the ImageNet Large Scale Visual Recognition Challenges (ILSVRC) \cite{ILSVRC15}. To further accentuate the complexity, \cite{sinitsineditable} introduces a highly challenging configuration, leveraging the Natural Adversarial Examples (NAE) dataset \cite{hendrycks2021natural}, consisting of $7500$ natural images notorious for their arduous classification nature, where pre-trained models exhibit correct prediction rates of less than $1\%$ for NAEs.

\paragraph{Image Inpainting} 
Image inpainting, also known as image completion, is the task of reconstructing missing regions in an image \cite{szeliski2022computer}. The problem is formalized as a regression over functions mapping pixel coordinates within $[0,1]^2$ to pixel values in $[0,1]$ (grayscale) or $[0,1]^2$ to pixel values in $[0,1]^3$ (RGB). This task has so far received less attention from the knowledge editing community \cite{garnelo2018conditional}, leveraging datasets that encompass both the MNIST dataset, once again serving as a rudimentary example, and the CelebFaces Attributes Dataset (CelebA) \cite{liu2015faceattributes} for more challenging scenarios. The CelebA dataset presents a more demanding scenario, offering over 200,000 celebrity images, each accompanied by 40 attribute annotations, making it a challenging and comprehensive dataset for exploration.

\begin{table*}[t]
\centering
\begin{tabular*}{\textwidth}{@{\extracolsep{\fill}}cccccc@{}}
\toprule
\textbf{Dataset} & \textbf{Tasks}             & \textbf{\# Examples} & \textbf{\# Classes} & \textbf{\# Channels} & \textbf{Resolution} \\ \midrule
MNIST            & Classification, Inpainting & 70k                  & 10                  & 1            & 28x28               \\
CIFAR-10         & Classification             & 60k                  & 10                  & 3             & 32x32               \\
CIFAR-100        & Classification             & 600k                 & 100                 & 3             & 32x32               \\
ImageNet           & Classification             & 1.2 M                & 1000+               & 3             & 224x224 +           \\
NAE              & Classification             & 7.5k                 & 200+                & 3             & 224x224 +           \\
CelebA           & Inpainting                 & 200k                 & -                   & 3             & 178x218             \\ \bottomrule
\end{tabular*}
\caption{Most important datasets used in Computer Vision for Knowledge editing. MNIST, CIFAR-10, CIFAR-100 are generally regarded as ``toy'' datasets while ImageNet, NAE, CelebA as more challenging testbeds.}
\label{tab:CVDatasets}
\end{table*}

\subsection{Natural Language Processing}
\label{sec:tasks-datasets:nlp}

Natural Language Processing (NLP) is also a broad field, concerned with giving computers the ability to process and understand human language \cite{eisenstein2019introduction}. 
Like in computer vision, in recent years researchers and practitioners in the field have leveraged the power of neural networks with many outstanding results \cite{otter2020survey, soltan2022alexatm, FitzGeraldATM}. With the recent paradigm shift from supervised learning to pre-training followed by fine-tuning \cite{wang2022pre}, and the trend towards larger and larger models \cite{zhao2023survey}, the ability to perform a cheap edit of a model instead of an expensive fine-tuning has motivated an intense interest from the knowledge editing community. Undoubtedly, within the NLP realm, the most widely targeted tasks for knowledge editing are Fact-Checking when dealing with classification, and (closed book) Question Answering for language generation. Some recent works also explore open Text Generation, editing of factual relations with had hoc datasets and also Document Classification. Table \ref{tab:NLPDatasets} provides an overview of the datasets commonly used for those tasks.

\paragraph{Fact-checking}
Fact-checking is the task of assessing whether claims made in written or spoken language are true, often addressed as a binary classification task \cite{guo2022survey}. In this setting, examples are natural language claims coupled with binary labels, even though occasionally a third neutral option is available. The claim $x^{(i)}$ is encoded with some tokenization scheme (or pre-trained word embedding) as a sequence of integers (or semantic vectors), while the label $y^{(i)}$ can take one of two values, positive or negative (optionally a third, aforementioned neutral value). One can then have a neural network predict this value with an explicit classification layer, or alternatively a language model producing special tokens (e.g., \textit{True}/\textit{False} or \textit{Supports}/\textit{Refutes}) for the considered truth values, when prompted with the claim under consideration.
The Fact Checking task has been considered particularly appealing for the knowledge editing community \cite{de2021editing, mitchellfast, mitchell2022memory, huangtransformer} for at least a couple of reasons. First, the recent phenomenal success of language models also highlighted their proneness to generate reasonable but factually wrong natural language text \cite{ortega2021shaking,ji2023survey}. This degrades the system performance and fails to meet user expectations in many real-world scenarios, leading to a great interest in the ability to mitigate these hallucinations. Furthermore, reasonable edit sets are fairly easy to obtain, e.g. randomly flipping the labels of claims from pre-existing datasets from Support to Refutes and vice versa. The most widely used datasets for this task are FEVER \cite{thorne2018fever} and VitaminC \cite{schuster-etal-2021-get}. Both of them are extracted from Wikipedia and annotated by human experts, to arrive at (\textit{evidence}, \textit{wikipage}, \textit{claim}, \textit{label}) tuples. In both cases, the label can be \textit{Supports}, \textit{Refutes} or \textit{Not Enough Info} depending on whether the evidence supports or not the claim. To construct proper editing datasets $\{(x_e^{(i)}, y_e^{(i)})_{i=1}^N\}$ out of them, \cite{de2021editing} (for FEVER) and \cite{mitchell2022memory} (for VitaminC) grouped facts based on the same pages, augmented each fact $x_e^{(i)}$ with some rephrases $\tilde{x}_e^{(i)}$, to assess generality, and randomly flipped the label $y_e^{(i)} \neq y^{(i)}$ in each group. 

\paragraph{Question Answering}
The task of training a model to provide a correct natural language answer to a given question is referred to as question answering; more specifically, \textit{closed-book} question answering is restricted to the case when the model is only fed with the question, and not a selection of possible answers, or supporting corpus from which an answer can be extracted \cite{roberts2020much}. In this case, both the input $x$ and the target label $y$ are natural language sequences, and we test the extent to which the model parameters implicitly store the required knowledge to provide correct answers. The most widely adopted dataset is surely the zero-shot Relation Extraction dataset (zsRE) \cite{levy2017zero}, used in particular by \cite{zhu2020modifying, de2021editing, mitchellfast, mitchell2022memory, meng2022mass, meng2022locating, huangtransformer, hartvigsen2022aging}. 
This task and datasets are particularly appealing for knowledge editing. Indeed, \textit{closed-book} question answering can benefit greatly from knowledge editing, as pipelines that solve the task usually leverage factual knowledge to answer questions; in the case of neural networks, this knowledge is acquired during training and implicitly stored in the networks' parameters, and it is unclear how to tweak these parameters to correct wrong or undesired answers, especially as the networks grow bigger in size. \cite{meng2022locating} hypothesize that this factual knowledge takes the form of (\textit{relation}, \textit{subject}, \textit{object}) triples, with intermediate layers acting as key-value storage units. This formalization lends itself nicely to the definition of an editing objective, rather than directly the open-ended natural language generation task. Furthermore, \cite{levy2017zero}\ demonstrates that it is possible to reduce relation extraction to the problem of answering simple reading comprehension questions and provided in their dataset multiple templates for each relation. For example, the triple (\textit{occupation}, $s$, $o$) can be naturally extracted by answering one of the following questions: \textit{What did} $s$ \textit{do for a living?}, \textit{What is} $s$ \textit{'s job?} \textit{What is the profession of} $s$\textit{?}. The subject $s$ can then be modified to create editing examples. Section \ref{sec:direct_knowlege_editing_problem} further discusses factual knowledge and how different works have modeled it for improving knowledge editing.
Besides zsRE, knowledge editing of models solving Question Answering has been studied leveraging also on additional datasets such as t-REX \cite{elsahar-etal-2018-rex} and Natural Questions (NQ) \cite{kwiatkowski-etal-2019-natural}. Finally, as a more challenging flavor of the same task with added counterfactual information \cite{meng2022locating} introduced a new dataset called Counteract.

\paragraph{Further NLP tasks}
Beside the two popular tasks outlined above, \cite{mitchellfast} tested editing text generation from autoregressive GTP-like models on a special version of WikiText-103 \cite{merity2016pointer}, where they are considering as prompts ($x_e$) passages sampled from WikiText itself and as edit targets ($y_e$) 10-token samples from a pre-trained distilGPT-2 model. This was for them a valid challenging editing setup since for their target model for editing a greedy 10-token prediction agrees with these edit targets for $<1\%$ of examples they extracted.
Finally, more recently, \cite{hartvigsen2022aging} tested their methodology on a novel task for knowledge editing using the SCOTUS dataset from \cite{chalkidis-etal-2022-fairlex}. The classification task is to categorize U.S. Supreme Court documents over multiple decades into $11$ topics. What makes this task interesting is that, over time, categorization rules change, so that label distributions shift. We note how this is off the shelf particularly realistic for knowledge editing as much of the world knowledge memorized by networks evolves over time just like those labels shifts in the dataset and the target of knowledge editing can be seen as keeping updated such world knowledge.

\begin{table*}[h]
\centering
\begin{tabular*}{\textwidth}{@{\extracolsep{\fill}}lllll@{}}
\toprule
\multicolumn{1}{c}{\textbf{Dataset}} & \multicolumn{1}{c}{\textbf{Tasks}} & \textbf{Format}            & \multicolumn{1}{c}{\textbf{\# Examples}} & \multicolumn{1}{c}{\textbf{\# Classes}} \\ \midrule
FEVER                                & Fact Checking                           & (evidence, wikipage, claim, label)             & 420k                                     & 3                                       \\
VitaminC                             & Fact Checking                           & (evidence, wikipage, claim, label)             & 450k                                     & 3                                       \\
zsRE                                 & Question Answering                      & (subject, relation, object) & 120M                                     & -                                       \\
T-REx                                & Question Answering                      & (subject, relation, object) & 11M                                      & -                                       \\
NQ                                   & Question Answering                      & (question, answer)         & 320k                                     &  -                                      \\
CounterFact                          & Question Answering                     & (subject, relation, true object, false object)          & 22k                                      & -           \\
Wikitext                             & Text Generation                         & tokens                     & 100M                                     & -                                       \\
SCOTUS                               & Document Classification                 & (date, text, label)        & 9.2k                                     & 11                                      \\ \bottomrule
\end{tabular*}
\caption{Most important datasets for knowledge editing used in NLP. We report characteristics of the original datasets even though often for knowledge editing ad hoc version, preprocessed to make editing meaningful are used.}
\label{tab:NLPDatasets}
\end{table*}

\subsection{Other Applications}
\label{sec:tasks-datasets:misc}

Even though the majority of works in the knowledge editing literature has focused on the Computer Vision and Natural Language Processing fields, as described above, the general nature of the editing problem yielded interesting results also in other fields, and will likely yield more in more diverse fields and applications in the years to come.
Among these, to the best of our knowledge, the most notable examples are applications in safety-critical scenarios and to graph neural networks; in the following we briefly review works from both.

\paragraph{Safety-critical Systems}




Safety-critical systems are those systems whose failure may lead to consequences that are determined to be unacceptable, such as significant damage to properties, the environment, or people \cite{knight2002safety}. Deep neural networks have grown in popularity over the past decade and are now being used in safety-critical domains such as self-driving cars \cite{gupta2021deep}, healthcare \cite{tekkecsin2019artificial} and aviation \cite{sridhar2020applications}. Clearly, in such critical scenarios, being able to find and correct unsafe neural network behavior becomes a crucial objective. This has motivated a line of research within the knowledge editing community, that so far has only touched the aviation domain, specifically the aircraft collision avoidance problem. The systems currently employed provide safety guarantees at the cost of being poorly data efficient, and efforts have been made to integrate neural networks into the pipeline \cite{julian2016, julian2019guaranteeing}. As a consequence, several subsequent works \cite{sotoudeh2021provable,fu2021sound,liang2023repairing} from the knowledge editing community have proposed approaches for fixing unsafe behavior of neural networks integrated in safety-critical pipelines.

Developing a robust collision avoidance algorithm that reliably prevents collision without alerting excessively is challenging due to sensor error and uncertainty in the future behavior of the aircraft. The Airborne Collision
Avoidance System X (ACAS X) \cite{kochenderfer2011robust} family of collision avoidance systems formulates the problem of collision avoidance as a partially observable Markov decision process. The variant for unmanned aircraft, ACAS Xu, uses dynamic programming (DP) to then find a solution in terms of resolution advisories that avoid collisions while minimizing disruptive alerts. The DP process makes use of a massive lookup table, that makes storage costly and certification time-consuming for certified avionics systems. Therefore, \cite{julian2016} propose using a deep neural network for
compressing the table without loss of performance, as measured by a set of safety and operational metrics.  
There are seven real-valued state variables that define an aircraft encounter, describing its geometry in terms of the two aircraft involved: (1) distance from ownship to intruder (2) angle to intruder relative to ownship heading
direction (3) heading angle of intruder relative to ownship
heading direction (4) speed of ownship 5) speed of intruder 6) time until loss of vertical separation (7) previous advisory action. There are then five possible horizontal maneuver \textit{advisories} that the system can produce: clear-of-conflict, or adjusting course by turning left or right at two fixed angles (hence 4 more possibilities). The state variables are usually discretized arriving at $\approx120$ million points, and the aforementioned lookup table associates scores to all pairs of 120 million states and five actions. This table is what makes up the ACAS Xu dataset: $\{(x^{(i)}, y^{(i)})_{i=1}^{N}\}$ with $N = 5\times 120$ million, where $x^{(i)}$ represents a discretized seven-dimensional state, and $y^{(i)} \in \mathbb{R}^5$ is the vector of scores associated to each of the five possible actions in that state. With 600 million floating-point numbers, the table requires
over 2 GB of storage. The task for the neural network is to regress this table, minimizing parametric knowledge required (i.e., number of parameters) and error with respect to the table. It is interesting to note that this is an atypical regression problem, since we aim for the guarantee that
the optimal advisory remains the same. When the difference between the scores of the first
and second-best advisories is relatively small, simple regression techniques (e.g., minimizing the Mean Squared Error) can lead to the network realizing a different strategy from that of the original table. This is reflected in the design of the loss function. 
The network or its further refinements \cite{julian2019guaranteeing} is then verified via tools such as \cite{wang2018formal, katz2017reluplex}, that are able to prove several input-output-based security properties, e.g., that a clear-of-conflict advisory will always be issued if the intruder is sufficiently far away, thus providing formal guarantees about DNN behavior. 
These properties are formalized as implications of the form $\forall x, x \in B \implies f_{\theta}(x) \in C$, where $f_{\theta}$ is the function approximator realized by the DNN with parameters $\theta$, $B$ is a bounded region of the input space and $C$ a bounded region of the output space. The ACAS Xu case study has so far been of great interest for the knowledge editing community since one such security property has been found to not be satisfied by the original network, exposing an input on which the network was inconsistent with
the lookup table ($\phi_8$ in \cite{katz2017reluplex}. This discrepancy would then be addressed by retraining the DNN, thus leading to the central question of knowledge editing: how to fix the network behavior on a limited set of points without affecting its behavior on unrelated points. Each work mentioned at the beginning of the paragraph has addressed this problem differently, but sharing the same setup: once a failing security property for a network is identified, and one is able to generate counter-examples, i.e., pairs $(x^{(i)}, y^{(i)})$ such that $x^{(i)} \in B$ but $y^{(i)} \notin C$, a certain strategy for defining candidate edit pairs can be formalized, i.e., how to assign a particular $\bar{y}^{(i)}$ to $y^{(i)}$. Then, usually a subset of this becomes the edit set, while the remaining portion is chosen as generality set; finally, a locality set is defined by points correctly classified by the network, i.e., input-output pairs for which the properties under scrutiny hold true. 

\paragraph{Graph Neural Networks}
Deep learning models have been particularly successful when dealing with signals such as speech, images, or video,
in which there is an underlying Euclidean structure; however, recently, there has been a growing interest in trying to apply learning on non-Euclidean geometric data, for instance represented in the form of graphs~\cite{bronstein2017geometric}. Graph Neural Networks (GNNs) learn node representations by applying shared permutation invariant functions over local neighborhoods of nodes in the graph \cite{bronstein2021geometric}. These representations can then be used for tasks like node classification. For instance, assigning a category to each paper in a citation graph \cite{wu2020comprehensive}. GNNs have achieved prominent results in learning features and topology of graph data, however, knowledge editing for GNNs is rarely explored, despite their widespread applicability; therefore, \cite{liu2023editable} propose a method to edit these models, restricting to the aforementioned node classification task. 

The task can be formalized as follows: let $G = (V, E)$ be an undirected graph with $V = (v_1,\dots, v_{|V|})$ and $E = (e_1, \dots, e_{|E|})$ being the set of nodes and edges, respectively. Given a feature space $\mathcal{X}$ (e.g., the space of real-valued $d$-dimensional feature vectors $\mathbb{R}^d$), a node feature tensor $X \in \mathcal{X}^{|V|}$ and a label space $\mathcal{Y}$, the goal is to learn a representation $h_{v}$ from which a label $y_v \in \mathcal{Y}$ for each node $v \in V$ can be easily predicted. Many datasets for node classification exist in the literature, comprised of data from various domains like citation networks and social networks, and we find again the distinction between small-scale and large-scale datasets \cite{wu2020comprehensive}. Among the former we find datasets like Cora, which contains a selection of Machine Learning papers collected from the web, and the references automatically parsed from their bibliographies \cite{mccallum2000automating}. In particular, the network contains $|V| = 2708$ nodes (articles) and $|E| = 5429$ links (citations); the feature space is $\mathcal{X} = \{0, 1\}^{1433}$, i.e., each node is described in terms of the presence (or absence) of certain keywords, taken from a dictionary of 1433 unique words (bag-of-words content representation); the label space is $\mathcal{Y} = \{1, \dots, 7\}$, i.e. the task is to predict to which of seven classes (e.g., \textit{Theory} or \textit{Reinforcement\_Learning}) each publication belongs. The Reddit dataset is instead a popular representative of large-scale node classification datasets. \cite{hamilton2017inductive}. constructed a graph dataset from Reddit posts made in the month of September 2014. The node label $y_v \in \mathcal{Y}$ in this case is the community, or “subreddit”, that a post belongs to, considering $|\mathcal{Y} = 50$ large communities. A post-to-post graph is constructed connecting posts if the same user comments on both. In total, this dataset contains $|V| = $ 232,965 posts with an average degree of 492 ($|E|$ is in the order of 100 million edges). In both of these cases, and the many other instances of node classification datasets, constructing an edit dataset is fairly straightforward, and done in the same manner as for the image classification task in Computer Vision. After training the model $f_0(\cdot)$ under consideration on a \textit{subgraph}, one evaluates it on the whole graph: each pair $(x^{(i)}, y^{(i)})$ such that $\hat{y}^{(i)} = \arg \max f(x^{(i)})$ is incorrect, becomes an edit pair $(x_e^{(i)}, y_e^{(i)})$. The geometry of graphs lends itself nicely to defining also generality and locality sets: indeed, since the task under consideration is node classification, as we have seen a single example $x^{(i)}$ describes a node $v$, then one can define its neighborhood $N(x^{(i)})$ to be its actual neighborhood in the graph $N_G(v) = \{ w \in V \mid \exists \ (v, w) \in E \}$; from this definition, generality, and locality sets follow consequently, as seen in earlier sections.

\section{Knowledge Editing Methodologies}
\label{sec:methodologies}
In recent times, several "knowledge editing" methods have been introduced to effectively modify the behaviors of models while maintaining their previous performance on the same task \cite{sinitsineditable}. These approaches draw inspiration from various fields of artificial intelligence research and can be broadly categorized into four distinct families: \textit{regularization techniques}, \textit{meta-learning}, \textit{direct model editing}, and \textit{architectural strategies}.

Regularization techniques utilize various forms of regularization to guide the model's learning process during fine-tuning, encouraging it to incorporate the desired edits while retaining its original capabilities \cite{zhu2020modifying}. Meta-learning approaches, on the other hand, employ hypernetwork models to learn parameter updates, enabling efficient adaptation to new tasks or knowledge \cite{de2021editing, mitchellfast}.
Direct model editing methods involve directly modifying the model's parameters or representations to incorporate the desired changes. These techniques can range from simple parameter updates to more complex approaches that leverage the model's internal representations. Finally, architectural strategies explore modifying the model's architecture itself, either by introducing new components or restructuring existing ones, to facilitate the integration of new knowledge or behaviors. In the upcoming sections, we will provide detailed discussions on each of these families of knowledge editing methodologies, highlighting their respective areas of application, advantages, limitations, and notable techniques within each category.

The objective of this section is to categorize various knowledge editing techniques discussed in the literature into the four distinct families mentioned above. All presented works have different characteristics, target different areas of application, types of edits, and adopt diverse experimentation strategies. Nevertheless, the objective of all the works reported can be formulated within the formal framework given in \ref{sec:knowlege_editing_problem}. A comparison between the most notable KE methodologies at the time of writing can be found in Table \ref{tab:comparison}. On the other hand, Table \ref{tab:zsRE&CounterFact} presents a comparison with non-sequential single batch edits of factual knowledge.

\subsection{Regularization Techniques}
\label{regularization_techniques}
\begin{table}[t]
\begin{tabular}{@{}lllccccccc@{}}
\toprule
    \textbf{KE Methodology} & 
    \begin{tabular}[c]{@{}l@{}}KE \\ Cathegory\end{tabular}  & 
    \begin{tabular}[c]{@{}l@{}}Training \\ Required\end{tabular}  & 
    \begin{tabular}[c]{@{}l@{}}Preserves\\ Architecture\end{tabular} & 
    \begin{tabular}[c]{@{}l@{}}Only\\ $(x_e, y_e)$\end{tabular} & 
    SNS & 
    BNS & 
    SSE & 
    BSE & 
    \begin{tabular}[c]{@{}l@{}}Scales\\ to LM\end{tabular} \\ 
    \midrule
        FT + $L_2$            & Regularization & False       & \greencheck   & \redxmark     & \greencheck   & \greencheck   & \greencheck   & \greencheck   & \redxmark     \\ 
        FT + KL                   & Regularization & False       & \greencheck   & \redxmark     & \greencheck   & \greencheck   & \greencheck   & \greencheck   & \redxmark     \\ 
        EWC                       & Regularization & False      & \greencheck   & \redxmark     & \greencheck   & \greencheck   & \greencheck   & \greencheck   & \redxmark     \\ 
        CNP                          & Architectural  & True       & \redxmark     & \greencheck   & \greencheck   & \greencheck   & \greencheck   & \greencheck   & \greencheck   \\
        ENN                          & Meta-Learning  & True       & \redxmark     & \greencheck   & \greencheck   & \redxmark     & \greencheck   & \redxmark     & \redxmark     \\ 
        KnowledgeEditor               & Meta-Learning & True & \greencheck   & \greencheck   & \greencheck   & \redxmark     & \redxmark     & \redxmark     & \greencheck   \\
        MEND                          & Meta-Learning & True  & \greencheck   & \greencheck   & \greencheck   & \greencheck   & \redxmark     & \redxmark     & \greencheck   \\ 
        MALMEN & Meta-Learning & True & \greencheck & \greencheck & \greencheck & \greencheck & \greencheck & \redxmark & \greencheck \\
        ROME                      & Direct Editing & False       & \greencheck   & \greencheck   & \redxmark     & \redxmark     & \redxmark     & \redxmark     & \greencheck   \\ 
        MEMIT                     & Direct Editing & False      & \greencheck   & \redxmark     & \greencheck   & \greencheck   & \greencheck   & \greencheck   & \greencheck   \\ 
        PMET & Direct Editing & False & \greencheck & \redxmark & \greencheck & \greencheck & \greencheck & \greencheck & \greencheck \\
        SERAC                         &  Architectural & True      & \redxmark     & \greencheck   & \greencheck   & \greencheck   & \greencheck   & \greencheck   & \greencheck   \\ 
        CaliNet       & Architectural &  True       & \greencheck     & \greencheck     & \greencheck   & \greencheck   & \redxmark   & \redxmark     & \greencheck   \\ 
        T-Patcher       & Architectural & False       & \redxmark     & \redxmark     & \greencheck   & \greencheck   & \greencheck   & \redxmark     & \greencheck   \\ 
        GRACE       & Architectural &  True       & \redxmark     & \greencheck     & \greencheck   & \greencheck   & \greencheck   & \greencheck     & \greencheck   \\ 
    \bottomrule \\
\end{tabular}
\caption{Comparison of the most notable KE methodologies in the literature. Different characteristics are reported for each approach, highlighting main advantages and disadvantages. For all approaches, we report: the category and whether it requires \textit{training} of an auxiliary model; if it \textit{preserves the architecture} of the edited model or requires the introduction of new components; whether it needs only the edit pair $(x_e, y_e)$, or requires additional input to perform the edit; if it is able to handle \textit{single non-successive} edits (SNS), \textit{batched non-successive} edits (BNS), \textit{single successive} edits (SSE), and \textit{batched successive} edits. Finally, we report if it can scale to Large Models (LM), that, following the definition in~\cite{zhao2023survey}, are models with more than 10B parameters.}
\label{tab:comparison}
\end{table}

Catastrophic forgetting, \cite{kemker2018measuring}, is a well-known phenomenon in literature, fundamentally limiting the flexibility of editing networks once trained or fine-tuned, \cite{lee2019overcoming}. Indeed, in the absence of any regularization process, the regular fine-tuning signal has the ability to easily execute a specific edit, albeit with a tendency to over-fit to the provided edit examples. However, this approach lacks in providing generality to the edit and has a negative impact on locality, owing to the absence of a plasticity mechanism, \cite{sinitsineditable}. Similar to continual learning, regularization techniques for knowledge editing aim to modify the standard fine-tuning signal of the target edit to ensure reliability and locality. Therefore, for regularization techniques, KE is not parametrized and does not require any pre-training, but it is nothing more than gradient descent computed with given edits and some specific regularization terms. While not all of these techniques were originally developed for the specific task of knowledge editing, they have proven to be in some way effective and are commonly used as useful baselines for comparison. Moreover, due to their simplicity, they can easily adapt to work with different types of edits: from single non-successive edits to batches of successive edits. However, depending on the number of layers fine-tuned, they rarely scale to models with large number of parameters such as Large Language Models (LLMs), that, according to \cite{zhao2023survey} are models with more than 10B parameters. In these cases, methodologies discussed in the following sections, may be better suited for efficiently editing large-scale with constrained resources.

For instance, authors in \cite{sotoudeh2021provable} introduce a technique called “Neural Network Patching” that allows for the correction of deep neural networks without retraining from scratch. They present the concept of neural network patching, which involves identifying and correcting the faulty or damaged components of a network. The proposed method involves using a set of repair operators to identify and replace damaged components of the network, identifying the minimum $L_{2}$ norm parameter update that reliably edits the model output, while minimizing the deviation from the original target model. Conversely, \cite{zhu2020modifying} utilize a constrained optimization process instead of penalizing an updated model for deviating from the original one. Their method employs either an $L_{2}$ or $L_{\infty}$ constraint between the original model parameters and the edited ones, highlighting the importance of selecting a subset of parameters to be updated. Additionally, they demonstrate how their method is similar to Elastic Weight Consolidation (EWC), \cite{lee2019overcoming}, which involves computing a penalty term for each weight in the neural network based on how much it contributes to the performance of the original task. Similarly to \cite{zhu2020modifying}, loss enriched with the Kullback–Leibler divergence (KL) have been proposed in order to regularize the network's weights based on the KL divergence between the network's output on the old task and the network's output on the new task. This encourages the network to maintain similar weights for the parts of the network that are relevant to both tasks \cite{yoon2017lifelong, serra2018overcoming, mitchellfast, huangtransformer}.

As previously mentioned, the techniques outlined in this section have been primarily utilized as baseline approaches in knowledge editing and heavily intersect with continual learning works \cite{mundt2023wholistic}. However, in various experiments cited in \cite{de2021editing, mitchellfast, huangtransformer}, these techniques have demonstrated limited efficacy in maintaining model accuracy with regard to previous knowledge. In fact, as noted by \cite{de2021editing}, regularization techniques overlook the highly nonlinear nature of large models and the significant impact that even minor changes in parameter space can have on the output of numerous data points. For those reasons, more elaborated works have been proposed in literature as the ones in the following sections.

\subsection{Meta-Learning and HyperNetworks}
\label{sec:meta-learning}

"Meta-learning techniques refer to a set of algorithms and approaches that enable machines to learn how to learn \cite{finn2017model}. These techniques have proven to be particularly useful in scenarios where a model needs to adapt quickly to new tasks or environments with limited data. For instance, in computer vision \cite{renmeta}, meta-learning can be employed to rapidly fine-tune a pre-trained model to recognize new object categories or adapt to different imaging conditions. Similarly, in natural language processing \cite{gu2020meta}, meta-learning can facilitate the adaptation of language models to different domains or styles of text with minimal additional training data. Robotics \cite{finn2017model} is another domain where meta-learning has shown promise, enabling robots to rapidly acquire new skills or adapt to changes in their environment. In reinforcement learning \cite{houthooft2018evolved}, meta-learning can help agents generalize their learned policies to new environments or task variations, improving their sample efficiency and adaptability.

The key advantage of meta-learning techniques lies in their ability to leverage prior knowledge and experience to learn new tasks or adapt to new environments more efficiently. By learning how to learn, these techniques can rapidly incorporate necessary modifications and adapt to new scenarios with significantly reduced data requirements, making them well-suited for applications where data is scarce, or the environment is dynamically changing.

They can be broadly categorized into two types: \textit{model-based} and \textit{optimization-based} meta-learning. In the context of knowledge editing, model-based meta-learning focuses on learning a model that can be used to adapt to new data efficiently. This involves learning the structure of the model and its parameters such that it can be generalized to new data. In optimization-based meta-learning, the focus is on learning how to optimize the parameters of a model to adapt to new knowledge.

\begin{table}[t]
\begin{tabular}{lcllllllll}
\toprule
Dataset                      & \multicolumn{1}{l}{Model} & \begin{tabular}[c]{@{}l@{}}Evaluation \\ Metrics\end{tabular} & FT    & KE    & MEND  & ROME  & MEMIT & SERAC & T-Patcher \\
\midrule
\multirow{3}{*}{zsRE}        & \multirow{6}{*}{GPT-J [6B]}    & SR                                                  & 54.70 & 6.60  & 45.60 & 99.18 & 99.23 & 90.16 & 97.12     \\
                             &                           & GR                                           & 49.20 & 7.80  & 48.00 & 94.90 & 87.16 & 89.96 & 94.95     \\
                             &                           & DD                                                      & 62.76 & 5.82  & 11.79 & 0     & 0     & 0.1   & 3.76      \\
\multirow{3}{*}{CounterFact} &                           & SR                                                  & 99.90 & 13.40 & 73.80 & 99.80 & 99.90 & 99.78 & 100.00    \\
                             &                           & GR                                          & 97.53 & 11.00 & 74.20 & 86.63 & 73.13 & 99.41 & 83.98     \\
                             &                           & DD                                                      & 98.98 & 5.62  & 96.25 & 0     & 0     & 1.11  & 91.63    \\
\bottomrule \\
\end{tabular}
\caption{Comparison of some of the most notable KE methodologies on zSRE \cite{levy2017zero} and ConterFact \cite{meng2022locating} with non-sequential single batch edits of factual knowledge. For each methodology, we report success rate (SR), generalization rate (GR), and Drawdown (DD) metrics. Results adapted from \cite{yao2023editing}.
}
\label{tab:zsRE&CounterFact}
\end{table}

In literature, authors of “Editable Neural Networks” (ENN), \cite{sinitsineditable}, firstly exploit the meta-learning paradigm for knowledge editing to “learn to allow effective editing”. The core idea behind “Editable Training” is to enforce the model parameters to be “prepared” for the editor function KE (Section \ref{sec:knowlege_editing_problem}), that in their experimentation is defined as Stochastic Gradient Descent (SGD) with up to $k$ steps and learning rate $\alpha$. In particular, they propose to train the starting model $f_{0}$ with a loss that at the same time encourages reliability and locality. That is obtained with two additional terms to the base loss, one that measures the success of an edit, cross-entropy, and another that assesses the distance in the output probability between the edited model and the original one, KL divergence. They prove to be successful on either computer vision datasets for classification \cite{krizhevsky2009cifar, deng2009imagenet} and Natural Adversarial Examples (NAE) \cite{hendrycks2021natural} and machine translation \cite{cettolo2014report}. They work with single non-successive and successive edits, with good performances on both types of edits. Nevertheless, as pointed out by \cite{mitchellfast}, the methodology requires to further training the base model before an edit with a pre-training step. That can be critical in scenarios with large models (LM) or when the memory available is a constraint.

Conversely, \cite{de2021editing} they propose an optimization-based meta-learning, firstly employing a hypernetwork \cite{ha2016hypernetworks} dubbed KnowleEditor with the objective to “learn to update” the parameters of another network. Therefore, since the knowledge editing task requires being locale except for the edit, they frame the learning task as a constrained optimization problem using a Bidirectional-LSTM \cite{schmidhuber1997long} and some downstream Feed-Forward Neural Networks (FFNN). Once trained, only the input edit feed the hypernetwork that predicts vectors to gate and shift the edit fine-tuning gradient of the starting network in respect to a certain weight matrix. Therefore, we can say that KnowledgeEditor learns how to modify the gradient in order to provide the property enumerated in Section \ref{sec:editing-properties}.
On the other hand, the training itself of the hypernetwork does not change the weights of the starting model as \cite{sinitsineditable}, but requires some original training samples to estimate the original output probability distribution. Indeed, as \cite{sinitsineditable}, the hypernetwork is trained with the sum of two loss components: the first one aims at providing reliability and generality with cross-entropy between semantically equivalent edits and predictions of the edited model and the second one provides locality with the KL divergence as \cite{sinitsineditable}. However, in addition to them, they propose to add a margin to KL and iteratively reduce it to progressively make the edit more local. They test their KnowledgeEditor on different NLP tasks, from fact-checking (FC) with a fine-tuned BERT model \cite{devlin2018bert} with the FEVER dataset \cite{thorne2018fever}, to closed book question answering (QA) with BART model \cite{lewis2020bart} on the Zero-Shot Relation Extraction (zsRE) dataset \cite{levy2017zero}. However, they only experimented with single non-successive changes. Nevertheless, authors of \cite{huangtransformer} adopt KnowledgeEditor as a baseline and their experiments on FEVER and zsRE show that it fails to implement more than a couple of successive single edits. Indeed, as hypothesized by the authors, KnowledgeEditor hypernetwork is trained with the starting model $f_{0}$ and thus strongly coupled with the original parameters. As the editing proceeds, the model becomes more different from the initial one, resulting in their failure.

Building over \cite{de2021editing}, authors of \cite{mitchellfast} leverage on hypernetworks too in order to learn how to update weights of a starting model. However, while KnowledgeEditor trains a recurrent neural network to map the edit example into a rank-1 mask over the gradient, \cite{mitchellfast} hypernetwork, named MEND, directly maps the gradient into a new parameter update, retaining tractability by leveraging the low-rank form of the gradient. Indeed, the input of a MEND network is a decomposed gradient and the output is formed by some pseudoactivations and psudodelta that should encapsulate, reliability, generality, and locality. Indeed, as \cite{de2021editing}, they make use of cross-entropy between semantically equivalent edits and predictions of the edited model to enforce generability and reliability (edit loss), and KL divergence for locality without the margin (locality loss). In addition, they propose two further hyperparameters to perform a weighted sum of the two losses and to make learnable the edit learning rate coefficient. Each layer of the network itself consists of two consecutive blocks, both initialized to compute the exact identity function of the normalized decomposed gradient, using four matrices initialized with zero or Xavier uniform initialization \cite{glorot2010understanding}. Finally, in order to edit multiple layers of the starting model with same matrices dimension, they propose to use the same MEND network, while applying a learned layer-specific scale and offset similar to \cite{perez2018film}. 
As \cite{de2021editing}, they experiment their methodology only with NLP tasks using FEVER \cite{thorne2018fever} and zsRE \cite{levy2017zero} datasets. In addition, they also evaluate with GPT-style models, \cite{radford2018improving}, working with a custom version of Wikitext-103 \cite{merity2016pointer}, named Wikitext Generation. They both experiment with non-sequential single or batch types of edits, showing large regression over 100 simultaneous edits. Finally, it is important to point out that as KnowledgeEditor, also MEND has the same limitations with successive edits, being strictly tied to the weights of the starting model.
In fact, if pre-training is conducted on $f_{t}$, KE will exhibit significantly poorer performance with $f_{t+1}$. Moreover, as the weights of the edited model diverge from those of the pre-training, MEND will gradually deteriorate, ultimately losing all its editing capabilities.

Similar to MEND, MALMEN \cite{tan2023massive} utilizes a hypernetwork to generate parameter shifts that can be applied to the language model. However, MALMEN distinguishes itself by formulating the parameter shift aggregation as a least squares problem, which it solves using the normal equation. This allows MALMEN to effectively combine the parameter shifts corresponding to different facts, mitigating the potential cancellation effects that can arise when simply summing the shifts. Additionally, MALMEN separates the computation between the hypernetwork and the language model, enabling the use of arbitrary batch sizes on both components. This memory-economic training strategy permits MALMEN to scale to editing thousands of facts simultaneously, substantially outperforming the editing capabilities of MEND. The paper provides thorough empirical evaluations demonstrating MALMEN's superior scalability and effectiveness across various language model architectures and knowledge-intensive NLP tasks, including closed book fact verification on the FEVER dataset \cite{thorne2018fever} and question answering on the zsRE dataset \cite{levy2017zero} for BERT-base, GPT-2, T5-XL (2.8B), and GPT-J (6B). Finally, it is important to point out that as KnowledgeEditor, and MEND, also MALMEN has the same limitations with successive edits, being strictly tied to the weights of the starting model.

\subsection{Direct Model Editing}
\label{sec:direct_knowlege_editing_problem}
\begin{figure*}[t]
    \centering
    \includegraphics[width=\textwidth]{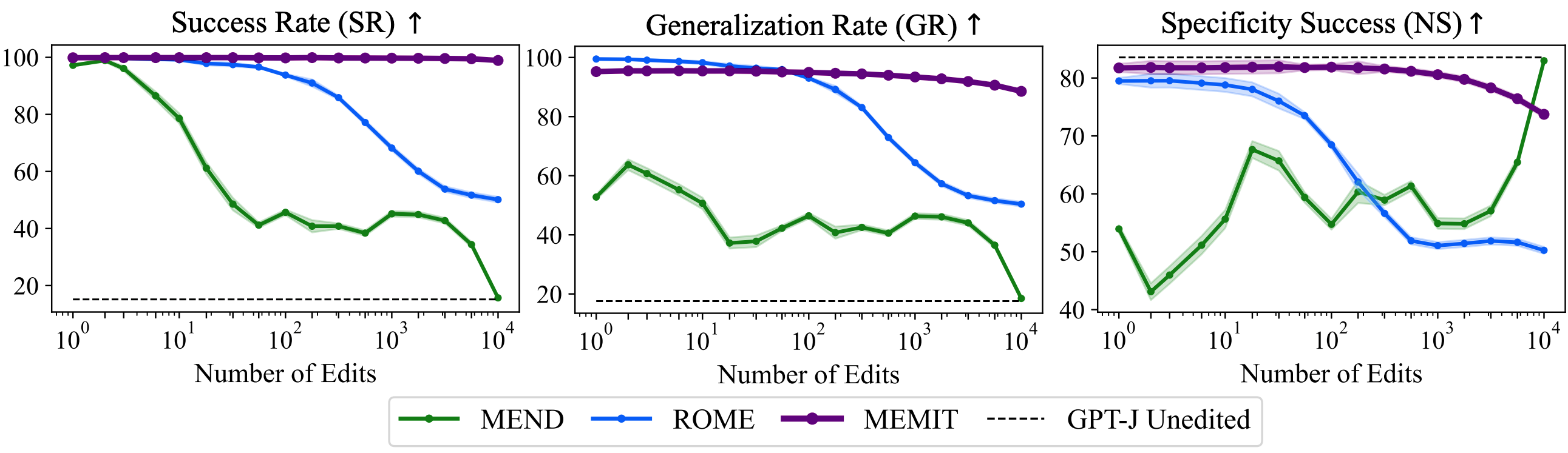}
    \caption{Scaling curves showing three different evaluation metrics with an increased number of non-successive batch edits for three different KE methodologies: MEND, ROME, and MEMIT. Results are computed using \textsc{CounterFact} and GPT-J. Locality is shown not as drawdown (DD), but as its complementary specificity over a neighborhood of samples \cite{meng2022locating}. ROME and MEND performs well up to ten edits, but rapidly degrade, losing almost all SR before batches of 1k. On the other hand, MEMIT performs well with considerable large batches of edits. Adapted from \cite{meng2022mass}.}
    \label{fig:methodology_comparison}
\end{figure*}
This family of approaches, known as direct model editing, aims to directly edit the weights or parameters within the target neural network, enabling effective implementation of a predefined set of changes while minimizing computational overhead. These methodologies allow injecting knowledge by modifying only certain parameters of the model, making them particularly suitable for foundational models
\cite{chang2023survey}. By directly manipulating the variables of a model, most direct model editing techniques build upon efforts to localize and understand the internal mechanisms within models, striving to attribute knowledge acquired during training to specific neurons or parameters in the network \cite{elhage2021mathematical, dar2023analyzing, mitchell2022memory}. Consequently, these approaches endeavor to edit the activations or values of identified neurons or parameters to reflect the desired changes.

However, it is crucial to recognize that the applicability and effectiveness of direct model editing techniques may vary depending on the underlying neural network architecture and the nature of the knowledge or changes to be incorporated. While extensive research has been conducted on specific architectures such as Multilayer Perceptrons (MLPs) and Transformers, the generalization of these techniques to alternative architectures or problem domains warrants further investigation. Therefore, an in-depth analysis and exploration of the applicable scenarios for direct model editing techniques are beneficial, considering factors such as the model architecture, the type of knowledge or changes to be incorporated, and the computational constraints of the target environment.

Geva, Mor, et al. \cite{tay2022transformer}, firstly identify the MLP layers in a masked LM transformer as key-value memories of entities and information associated with that entity, \cite{sukhbaatar2015end}. Building  on this finding, \cite{dai2022knowledge} demonstrate a method to edit facts in BERT models \cite{devlin2018bert}; they propose the concept of knowledge neurons to study how factual knowledge is stored in pretrained Transformers. Specifically, the authors examine the fill-in-the-blank cloze task for BERT and propose a knowledge attribution method to identify the neurons that express a given relational fact. They find that the activation of such knowledge neurons is positively correlated to the expression of their corresponding facts. Therefore, they present a preliminary methodology that leverages knowledge neurons to edit factual knowledge in Transformers, even without any fine-tuning. The authors perform a knowledge surgery for pretrained Transformers by directly modifying the corresponding parameters in feed-forward networks. Such surgery shows promising results, keeping a moderate influence on other knowledge. The methodology proposed can be used to perform both single and multiple edits at the same time. However, the authors only experimented with single edits in their paper, focusing on specifically on factual knowledge. 

Factual knowledge is also the focus of Meng Kevin, et al. work, \cite{meng2022locating}, but exclusively experimenting on autoregressive language modelling, such as GPT-style models, \cite{radford2018improving}. The study explores the storage and direct manipulation of factual associations in transformer models. These associations are modeled as tuples, represented as $t=(s, r, o)$, comprising a subject $s$, a relation $r$, and an object $o$, which connect to form a factual relationship. The research focuses into understanding how these tuples are stored within transformer models and investigates methods to facilitate their direct editing.
 Firstly, they trace the causal effects of hidden state activations within GPT using causal mediation analysis \cite{pearl2022direct} as previously done by other works \cite{vig2020investigating} to identify the specific modules that mediate recall of a fact about a subject. Their analysis reveals that feedforward MLPs at a range of middle layers are decisive when processing the last token of the subject name. Secondly, they test their previous findings by introducing a novel direct model editing methodology: Rank-One Model Editing method (ROME). The algorithm aims at producing single, not successive edits by altering the parameters that determine a feedforward layer's behavior. As stated by the authors, despite the simplicity of the intervention, ROME is similarly effective to other model-editing approaches achieving good generalization and locality simultaneously, whereas previous approaches sacrifice one or the other. Furthermore, they introduce a new dataset, CounterFact (derived from a set of true facts from WikiData), for evaluating counterfactual edits in language models. It contains 21,919 records with a diverse set of subjects, relations, and linguistic variations. Finally, they propose to monitor a further metrics, \textit{fluency}, to evaluate text generation's repetitiveness by measuring the weighted average of bi- and tri-gram entropies \cite{zhang2018generating}.

 ROME results are comparable with other knowledge editing meta-learning techniques. Nevertheless, the methodology is limited to work with single non-successive edits. To address this problem, the same authors modified the original algorithm to overcome this problem \cite{meng2022mass}. The method, called MEMIT, is able to scale up to thousands of associations for GPT-J (6B) and GPT-NeoX (20B), exceeding prior work by an order of magnitude. As ROME, MEMIT works by first identifying the critical layers in the LLM that are responsible for storing the memories that need to be updated. Then, it uses a rank-one modification of the MLP weights of these layers to directly write the new memories into the model. This process is repeated for each memory that needs to be updated. The authors evaluated MEMIT on a variety of tasks, including factual question answering, natural language inference, and summarization. They found that MEMIT was able to significantly improve the performance of the LLM on these tasks, even when the number of memories that were updated was large. In addition to its scalability, MEMIT also has several other advantages over previous methods for updating LLMs. First, it is relatively efficient, requiring only a few minutes to update a large model with thousands of memories. Second, it is very accurate, with the updated models achieving near-perfect accuracy on the tasks that were evaluated. Third, MEMIT is generalizable, with the updated models able to perform well on a variety of tasks. Figure \ref{fig:methodology_comparison} compares ROME, MEMIT, and one of the meta-learning techniques described in section \ref{sec:meta-learning}. It is possible to appreciate the improvement brought with MEMIT and the clear collapse of the MEND HyperNetwork-based editor with large batches. Overall, MEMIT is a valuable tool for improving the performance of these models. However, it is important to point out that, even if there are works that try to expand the scope of the methodology \cite{gupta2023editing}, MEMIT is limited to modifying factual association and that there is not a clear path to scale it to different knowledge types. The same applies for PMET \cite{li2024pmet} 
which goes a step further by also simultaneously optimizing the hidden states of the Multi-Head Self-Attention (MHSA) component. This is based on the insight that MHSA encodes certain general knowledge extraction patterns that can be leveraged to enable more precise editing of the model. On the other hand, it is worth noticing that recent research suggests that casual tracing or other knowledge localization methodologies for identifying which parameters to update are surprisingly unreliable also for factual knowledge \cite{hase2023does}. The paper argues that locating the source of a fact does not necessarily translate to effective editing, as modifying that specific location may not change the model's output. This surprising finding indicates our understanding of how knowledge is stored in complex models is incomplete.

\subsection{Architectural Strategies}
Architecture strategies represent a distinct family of methodologies that diverge from approaches that directly or indirectly modify the weights of the target model. Instead, these strategies concentrate on manipulating or augmenting the original architecture to patch and modify the network's existing knowledge. This approach can be advantageous in scenarios where the set of original weights is inaccessible or when it is deemed unsafe to directly manipulate the main model. This indirect manipulation can be particularly useful when dealing with highly sensitive or proprietary models, where direct weight modification may be prohibited or could potentially compromise the model's integrity. Additionally, architecture strategies can offer a more interpretable and controllable means of knowledge adaptation, as the introduced architectural changes can often be mapped to specific functional aspects of the model. However, it is always crucial to consider the trade-offs between the potential benefits of architectural manipulation and the computational overhead associated with it in terms of latency.

A first editor function of this type was proposed by \cite{sinitsineditable}, that taking inspiration from Conditional Neural Processes (CNP) \cite{garnelo2018conditional}, they propose a specialized architecture that performs edits by adding a special condition vector to intermediate activations. The vector is generated by an additional encoder layer and provides information to effectively steer the final prediction of the network. This approach is very similar to more generic memory-augmented models that introduce memory mechanisms to enhance neural networks \cite{graves2014neural, santoro2016meta}. On the same line of experimentation, Mitchell et al. \cite{mitchell2022memory} proposed an approach, called SERAC (Semi-Parametric Editing with a Retrieval-Augmented Counterfactual Model), which stores edits in an explicit memory and learns to reason over them to modulate the base model's predictions as needed. This allows SERAC to be more expressive than other model editors, and to produce more accurate predictions for test inputs that are not closely related to the edit. That is achieved using three components: a memory, a classifier, and a conterfactual model. Users adds edit in the memory and the classifier decides whether the memory contains inputs relevant to processing them. If the classifier determines that a relevant edit example exists, the input and edit example are passed to the counterfactual model that is responsible for making the prediction. They evaluate SERAC on numerous tasks, focusing exclusively on large language models with single and batch non-successive edits.

On the same line of SERAC, authors of \cite{dong2022calibrating} presented a memory-based methodology called CaliNet. It is a method for calibrating specifically factual knowledge in pretrained language models (PLM) without re-training from scratch. CaliNet first detects whether a PLM has learned a fact correctly by comparing its scores for the correct fact and a set of distractor facts. If the PLM has not learned the fact correctly, CaliNet then uses a lightweight method to add and adapt new parameters to the PLM for that specific fact. CaliNet has two main components: first, a contrastive knowledge assessment (CKA) module that detects whether a PLM has learned a fact correctly. The CKA module works by comparing the PLM's scores for the correct fact and a set of distractor facts. If the PLM assigns a higher score to the correct fact than to the distractor facts, then the CKA module concludes that the PLM has learned the fact correctly. Secondly, a factual knowledge calibration (FKC) module adds and adapts new parameters to the PLM for a specific fact. The FKC module works by first creating a new set of parameters for the PLM that are specific to the given fact. These new parameters are then initialized with the PLM's original parameters for the fact. The FKC module then fine-tunes the new parameters on a dataset of factual examples that include the given fact. Using a custom dataset based on ParaRel set \cite{elazar2021measuring}, CaliNet has shown to be effective at calibrating factual knowledge in PLMs. In experiments on the knowledge probing task, CaliNet was able to improve the factual accuracy of PLMs by up to 20\%.

All previous works have tackled only non-successive edits, making only one edit or batch of edits at a time. However, as pointed out by \cite{huangtransformer}, the one-mistake-fixing scenario is not an accurate abstraction of the real-world challenge. Therefore, they extend the scenario into successive edits, introducing a novel model editor, Transformer-Patcher (T-Patcher), that can shift the behavior of transformer-based models by simply adding and training a few neurons in the last Feed-Forward Network layer. Being at the end of the model, the new neurons have access to the output of all the previous layers. This allows the new neurons to learn how to correct mistakes that are made in earlier layers of the model. Overall, the proposed methodology allows for targeted shifts in the behavior of the model, akin to other fine-tuned oriented methodologies like LoRA \cite{hu2022lora}.
Transformer-Patcher is able to successively correct up to thousands of errors while generalizing to their equivalent inputs and retaining the model's accuracy on irrelevant inputs. This is in contrast to previous methods, which either fail to make a sequence of edits or to remember previous edits. The work evaluates Transformer-Patcher on both classification and generation tasks, showing that Transformer-Patcher can achieve state-of-the-art performance for single successive edits. However, despite the expansive scope of the methodology, their approach and implementation are highly architecture-specific, relying heavily also on large sources of data and unrelated inputs.

To this end, Hartvigsen, Thomas, et al. \cite{hartvigsen2022aging}, proposes GRACE, a methodology that requires only samples of the edits and that can perform successive edits, ensuring minimal impact on unrelated inputs. GRACE works by caching a chosen layer's activations in an adaptive codebook as edits stream in. When a new edit is received, GRACE retrieves the corresponding activations from the codebook and uses them to update the model's predictions. This allows GRACE to make targeted edits to the model's behavior without affecting its overall performance. The authors evaluated GRACE on a variety of tasks, including text classification, natural language inference, and question answering. They showed that GRACE was able to significantly improve the performance of LLMs on streaming errors, while minimally affecting their performance on unrelated inputs. 
Finally, the authors do not explicitly mention the impact of GRACE on latency. However, it is possible that GRACE could have a small impact on latency, as it requires caching activations in an adaptive codebook.

\section{Conclusion and Future Directions}
\label{sec:conclusion}
In this survey, we organize recent progress of the increasingly growing field of knowledge editing. We began formalizing it under a common definition that seamlessly connects the slightly different facets of each work presented in the literature so far. Indeed, being a very recent branch of research, each author tried to differently bend definitions to better accommodate their methodologies. So, following the first definitions given by \cite{sinitsineditable} and the more recent adaptation by \cite{huangtransformer}, we formally define knowledge editing as the task of modifying the knowledge of a target model in a non-sequential or sequential manner, without significantly affecting its original performance. This makes knowledge editing closely related to the much more well-known branch of continuous learning, or lifelong learning, as well as the emerging fields of machine unlearning \cite{bourtoule2021machine, nguyen2022survey, si2023knowledge} and parameter-efficient fine-tuning \cite{fu2023effectiveness, hu2022lora, han2024parameter, yu2024melo}, including most notable variants like model editing via task arithmetic \cite{ilharco2022editing, hendel2023context}. In the following section, we briefly discuss the intersections and connections between knowledge editing and related disciplines.

\subsection{Knowledge editing and related fields of research}

Machine unlearning researches focus on removing specific data samples or knowledge from a pre-trained model without retraining the entire model from scratch. This task is gaining relevance due to its applications in data privacy \cite{chen2021machine}, security \cite{cao2015towards}, and compliance with regulations such as GDPR \cite{sai2023machine}. However, while knowledge editing primarily focuses on modifying or adding knowledge to a model, machine unlearning can be seen as a complementary task that aims to selectively remove knowledge from a model. Both tasks share the common goal of avoiding complete retraining while maintaining the model's predictive capability, but tracking different metrics and evaluation benchmarks.

Knowledge editing and machine unlearning can be viewed as specialized cases of continual learning \cite{mundt2023wholistic}, focusing on applying targeted, often non-uniformly distributed edits to a network's existing knowledge \cite{henn2021principled}. In contrast, continual learning encompasses a broader scope, seeking general methodologies to incrementally expand a network's knowledge base, enabling it to perform an increasing array of tasks and skills over time. The key distinction lies in their objectives: knowledge editing aims for precise modifications to specific knowledge elements, while continual learning emphasizes the gradual accumulation of knowledge and skills without compromising previously learned information. Despite these differences, both fields share significant challenges, particularly in modifying neural networks without disrupting existing capabilities. Both contend with issues such as catastrophic forgetting \cite{ratcliff1990connectionist}, where new learning can overwrite previously acquired knowledge. However, the approaches to addressing these challenges often diverge. Knowledge editing techniques typically employ localized, precise modifications, while continual learning methods often utilize more global strategies to balance retention of old knowledge with acquisition of new information. The intersection of these fields is particularly evident in scenarios where techniques from one inspire developments in the other. For example, some knowledge editing approaches adapt continual learning strategies for mitigating catastrophic forgetting (Section \ref{regularization_techniques}), tailoring them for more targeted edits. Conversely, insights from knowledge editing about precise neural network modifications can inform the development of more fine-grained continual learning algorithms.

On the other hand, parameter-efficient fine-tuning (PET) techniques are another related field that involves modifying pre-trained models to create new models with desired capabilities. PET aims to enable efficient adaptation by updating only a minimal subset of model parameters, rather than fine-tuning all parameters. Notable PET techniques include addition-based methods like adapters \cite{houlsby2019parameter}, specification-based methods like LoRA \cite{hu2022lora}, and reparameterization-based methods. A interesting variant of PET is task arithmetic \cite{ilharco2022editing}, which involves operations such as model addition, subtraction, multiplication, and permutation to modify or reshape the knowledge encoded in pre-trained models. By manipulating entire model representations in this arithmetic fashion, task arithmetic provides a way to create tailored models for specific tasks or domains without retraining from scratch, similar to knowledge editing techniques. However, task arithmetic and PET more in general are typically applied to enhance task performance rather than edit knowledge specifically. The efficacy of existing PET methods for knowledge editing remains largely unexplored. Indeed, while task arithmetic and other PET techniques hold promise for efficient model adaptation, they differ from knowledge editing in their primary focus on improving downstream task metrics rather than directly modifying a model's acquired knowledge. Investigating how to leverage PET methods for precise and targeted knowledge updates presents an interesting direction for future work in the knowledge editing space.

\subsection{Future Directions and Risks}

Building on the more formal definition, we presented a distilled summary of the most relevant works in the literature at the time of writing. We proposed to categorize these works into four families: regularization techniques, meta-learning, direct model editing, and architectural strategies. We discussed each family in turn, highlighting their intrinsic characteristics and limitations. We also summarized the most frequent field of application, tasks, and datasets that have been tackled in each family, for quick reference. We did not specifically deep dive into the works where knowledge editing could emerge as an additional benefit of the proposed methodologies \cite{hewitt2023backpack} but it is worth noting that future expansions of similar survey research could encompass these aspects for a more comprehensive analysis.

Overall, we have presented a rapidly expanding field of research driven by the current trend of foundational models \cite{zhou2023comprehensive}. The advancements in this area have led to a significant increase in the development of tools and methodologies to effectively harness the intrinsic knowledge of these models. As we move forward, knowledge editing is poised to become a critical factor in leveraging the power of these models for various industrial applications. However, several key challenges and future directions remain to be addressed:

\begin{itemize} 
\item Improving the efficiency and scalability of knowledge editing techniques to handle large-scale foundational models with constrained computational resources. This challenge is particularly pressing as models continue to grow in size and complexity \cite{dubey2024llama}. Future research could explore techniques such as sparse editing or hierarchical knowledge representations to enable more efficient updates.
\item Enhancing methodologies to introspect the different types of distributed knowledge stored inside neural networks, to enable better understanding and control over the changes made to models \cite{templeton2024scaling}. This could involve developing novel visualization techniques or interpretability methods specifically tailored for knowledge editing tasks.
\item Developing robust benchmarks specifically tailored for knowledge cross-domains editing tasks, to better assess the performance and impact of different techniques. These benchmarks should not only measure the effectiveness of edits but also their impact on the model's overall performance and generalization capabilities.
\item Investigating the potential of knowledge editing for improving the robustness, fairness, and ethical behavior of AI models, by correcting biases, removing harmful tendencies, or incorporating desired values and principles. This direction could lead to the development of "ethical editing" frameworks that ensure AI systems align with human values and societal norms.
\item Developing knowledge editing techniques to incorporate edits at runtime, allowing to create systems able to quickly adapt to errors and changes. This could pave the way for more adaptive and flexible AI systems that can learn and update their knowledge on-the-fly.
\item Investigating the potential of knowledge editing in multi-modal models \cite{baltruvsaitis2018multimodal}, where information is represented across different modalities (e.g., text, images, audio). This research direction could lead to more versatile and comprehensive knowledge editing techniques applicable to a wider range of AI systems.
\end{itemize}

Finally, it is vital to acknowledge that the power of knowledge editing also brings inherent risks that must not be overlooked. While editing models can correct their behavior and improve their utility, it can also be exploited for harmful purposes. In particular, sophisticated editing algorithms may enable malicious actors to deliberately incorporate backdoors, vulnerabilities, hidden behaviors, or harmful tendencies into the models. This concern becomes even more critical for methodologies that edit weights without providing sufficient interpretability of the applied changes. This dual use is a common risk for many machine learning technologies, and only proactive efforts to develop robust security measures and ethical guidelines can help to mitigate these potential risks. Future research should focus on developing "edit verification" techniques that can detect and prevent malicious edits, as well as establishing standardized protocols for auditing and certifying edited models.

In conclusion, knowledge editing in AI has emerged as a critical field, offering transformative potential for enhancing AI capabilities while simultaneously raising significant challenges and ethical considerations. As researchers and practitioners, we must strive to balance the pursuit of technological advancements with a strong awareness of their broader societal impacts, ensuring that the evolution of knowledge editing techniques contributes positively to the development of safe, reliable, and beneficial AI systems.

\bibliographystyle{unsrtnat}
\bibliography{references}  






\end{document}